\documentclass[11pt]{article}

\usepackage[final]{acl}

\usepackage{times}
\usepackage{latexsym}

\usepackage[T1]{fontenc}

\usepackage[utf8]{inputenc}

\usepackage{microtype}
\usepackage{subcaption}
\usepackage[table,xcdraw]{xcolor}
\usepackage{csquotes}
\usepackage{url}
\usepackage{tabularx}
\usepackage{booktabs}
\usepackage{paralist}

\usepackage{inconsolata}

\usepackage{graphicx}

\newcommand\changed[1]{#1} 

\title{Finding Sense in Nonsense with Generated Contexts: \\ Perspectives from Humans and Language Models}

\author{Katrina Olsen \\
  Grid Dynamics\thanks{This research was preformed while at IMS.} \\
  \texttt{kolsen@griddynamics.com} \\\And
  Sebastian Padó \\
  IMS, University of Stuttgart, Germany \\
  \texttt{pado@ims.uni-stuttgart.de} \\}

\begin{document}
\maketitle
\begin{abstract}
\changed{A}nomalous sentences have been instrumental in the development of computational models of semantic interpretation. \changed{Their detection is also central for the LLM lifecycle in tasks such as hallucination detection or automatic text cleaning. A} core challenge is \changed{distinguishing} what is merely anomalous (but can be interpreted given a supporting context) and what is \changed{purely} nonsensical. However, it is  unclear (a) how nonsensical, rather than merely anomalous, existing datasets of \changed{semantically unacceptable} sentences are; and (b) how well LLMs can make this distinction. In this paper, we answer both questions by collecting sensicality judgments from human raters and LLMs on sentences from five semantically deviant datasets---both context-free and when providing a context. We find that raters consider most sentences at most anomalous, and only a few as properly nonsensical. We also show that LLMs are substantially skilled in generating plausible contexts for anomalous cases.


\end{abstract}

\section{Introduction}

Nonsensical  sentences are often used in semantics research to probe the capabilities -- and limits -- of semantic interpretation models. For instance, models are asked to interpret  \enquote{jabberwocky} sentences with made-up words \cite{first-jabberwocky,prob-sym}, detect word salad sentences \cite{bert-bad-nonsense}, and answer incomprehensible questions reduced based off a model's attention \cite{og-path}. Semantically invalid sentences are also used as negative samples in plausibility and acceptability scoring \cite{neural-cola,surprised-anomalies,plaus-adept}.

Indeed, the wide variety ways a sentence can lack meaning has been cast into a range of semantically deviant and nonsensical datasets (cf. Table~\ref{tab:premade}). Some contain sentences whose components cannot come together into a consistent meaning \cite[]{eva-first,spicy,topic-anomaly}. 
Or, as explained by \citet{spicy}, a phrase may be meaningless by having too many possible interpretations, e.g., \enquote{residential steak.} 
This can be due to what \citet{ling-terminology} calls \enquote{pragmatic infelicity} or what \citet{spicy} lists as selectional restrictions and thematic fit. \citet{ling-terminology} include necessary falsehoods as nonsense, in that they cannot be truly meaningful in any situation.

\changed{This distinction also has substantial practical importance for data screening, where acceptability-based methods are often used in data cleaning and hallucination detection. False positives in this process can mean that unusual, yet valid, data is being discarded. The removal of such cases from training data can weaken model performance, as training on ambiguous, harder-to-classify data has been shown to contribute the most to out-of-distribution generalization 
\citep{dataset-mapping,wanli-dataset-mapping}. 
We therefore analyze how human and language model handle}
two subtypes of semantically deviant data. 
The first, which we call \enquote{anomalous,} is where sentences may be odd but still have the potential to be interpreted --- 
such as with creative metaphors, e.g., \enquote{The building bounced through the party.} when at a costume party \citep{metaphor-cusp}.
The second, which we call \textquote{pure nonsense}, are sentences that do not afford any interpretation due to true semantic violations, e.g., \enquote{settling dog transparent.} Models that over-rely on likelihood memorization rather than semantic parsing of interpretations will incorrectly treat both types as pure nonsense \cite{comp-gap}.

\begin{table*}[tb!h]
\begin{tabular}{llllp{8cm}}
\toprule
\textbf{Name} & \textbf{Lines} & \textbf{Type} & \textbf{Subtypes} & \textbf{Example} 
\\ \midrule

ADEPT & 4,086                                                       & Plausibility  &                                                              & Dead ice is melted.                                                                                   \\
\rowcolor[HTML]{EFEFEF} 
BLiMP & 2,000                                                       & Animacy       & \begin{tabular}[c]{@{}l@{}}Passive\\ Transitive\end{tabular} & \begin{tabular}[c]{@{}l@{}}Tina is fired by the radius.\\ Every pie knows Eva.\end{tabular}           \\
PAP   & 871                                                         & Plausibility  & \begin{tabular}[c]{@{}l@{}}Abstract\\ Concrete\end{tabular}  & \begin{tabular}[c]{@{}l@{}}The humor requires the merger.\\ The moon revolves the album.\end{tabular} \\
\rowcolor[HTML]{EFEFEF} 
CConS & 331                                                         & Relative Size & \begin{tabular}[c]{@{}c@{}}Explicit\\ Implicit\end{tabular}  & \begin{tabular}[c]{@{}l@{}}A chicken lived in a bird's beak.\\ A plane entered a car.\end{tabular}    \\
Cusp  & 200                                                         & Figurative    &                                                              & The date was a bus.                               \\             \bottomrule                                     
\end{tabular}
\caption{Datasets used in this study}
    \label{tab:premade}
\end{table*}

\begin{figure}[tb!]
    \centering
    \includegraphics[width=\columnwidth]{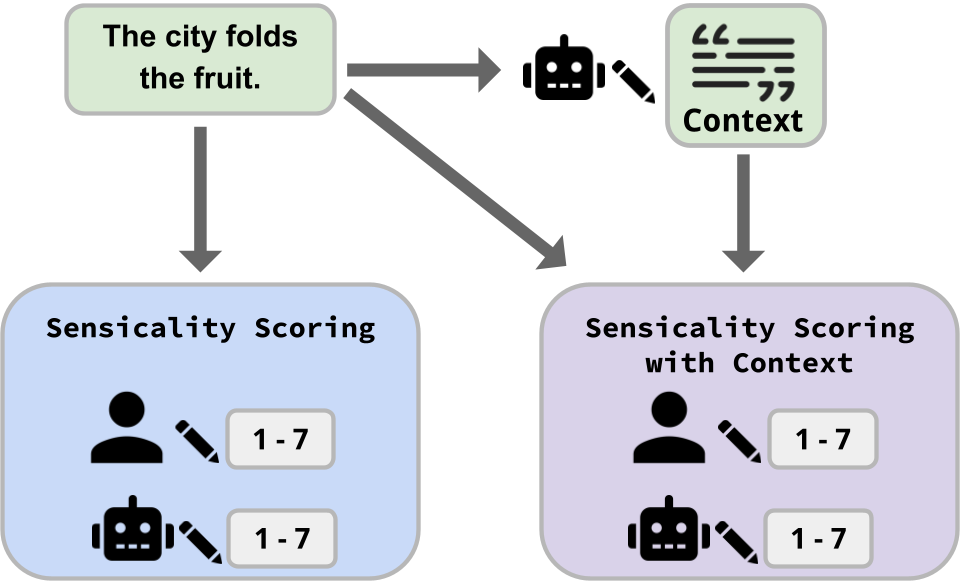}
    \caption{Visualization of sentence scoring process}
    \label{fig:methodology}
\end{figure}

In sum, we investigate two questions which, to our knowledge, have not received much attention:
\begin{enumerate}
    \item[RQ1.] \textbf{How much pure nonsense is in semantically deviant datasets?}
    Our hypothesis is that existing datasets covering the space of, broadly speaking, ``nonsensical'' sentences mix pure nonsense with semantic anomalies; but what is the predominance, and is it consistent across datasets?
    \item[RQ2.] \textbf{How well are 2025 SOTA models able to deal with nonsense?} 
    \changed{Given how LLMs supposedly have greater compositional generalization than  earlier models \citep{doi:10.1073/pnas.2417182122}, we consider it valuable to analyze their performance across a range of pre-existing semantically deviant datasets.}
\end{enumerate}

 We propose to operationalize the difference between pure nonsense and semantic anomalies through the generation of \textit{prior contexts}, anchoring a text's semantic interpretation to the pragmatic context in which it could be used. Contextualization can make semantic anomalies interpretable, but cannot do that for pure nonsense. %
Figure~\ref{fig:methodology} sketches our procedure, where human raters and LLMs judge the sensicality of semantically deviant sentences  both without context and in context.

Our results are: (RQ1) semantically deviant datasets contain many sentences that are fully sensical even without context, and little pure nonsense; and (RQ2) LLMs deal fairly well with deviant sentences, for both scoring and context generation.

\section{Related Work}

\changed{It is not safe to assume that LLMs have a strong capability, like human speakers,} to ground unusual sentences’ meaning in hypothetical contexts. An example is provided by the findings in \citet{counter-common}, where LLMs are unable to infer an object’s relative size when given unusual, seemingly impossible contexts. 
This aligns with what \citet{comp-gap} calls the “compositionality gap”, where models can succeed in most tasks but also struggle to process the semantic compositionality of unique phrases which had not been in their pre-training data. 
\citet{spicy} propose that, to test if distributionally-based models learn compositional meaning, one can evaluate a model’s ability to distinguish unattested but acceptable data from unattested, semantically deviant data \cite{eva-first,spicy}.
\changed{This is supported by \citet{drivel}, who use pragmatic paradoxes that they dub as \enquote{nonsense with depth} to probe LLM reasoning across tasks, revealing a consistent gap between LLMs' fluency and their genuine comprehension.}
\changed{Such findings} call into question whether LLMs are able to hypothesize beyond common scenarios to generate situations where a seemingly nonsensical sentence could be meaningful.

Despite the previously mentioned substantial body of linguistic research on nonsense, little work has been done on contextualized nonsense.
Beyond \citet{llm-reply-to-nonsense} detecting purely pragmatic-level nonsense in dialogue models, the majority of research in acceptability rating and semantic deviance detection is on individual sentences without context \cite{neural-cola,bert-bad-nonsense,metaphor-cusp}. 
An exception is \citet{accpt-in-context}, who investigate the effect of context on sentence acceptability judgments.


\section{Data}\label{preex}

For our context generation and human evaluations, we create a diverse sample across the spectrum of nonsensicality and impossibility by selecting a random 40 sentences from five of the most widely datasets on ``unusual'' sentences (cf. Table \ref{tab:premade}). When subtypes are available in a dataset, we select 20 sentences from each subtype.

\paragraph{ADEPT}
The ADEPT dataset was made for discerning 
the change in a sentence's plausibility when inserting an adjectival modification on a noun \citep{plaus-adept}. 
Each sentence is scored on a 5-point scale for how the inserted adjective affects the plausibility of the original sentence, from which we select sentences with the lowest plausibility, \enquote{impossible.}

\paragraph{BLiMP}
We utilize the \enquote{animate subject passive} and \enquote{animate subject trans} subtypes from the BLiMP dataset, which focuses on English grammatical errors \citep{blimp}. 
Both subtypes are cases where an non-animate subject is being used as animate, which we consider similar to the selectional preference violations explored in \citet{spicy}.

\paragraph{PAP}
\citet{pap} introduces a plausibility dataset of subject-verb-object triplets, labeled by each component's concreteness \citep{brys-concrete}. 
We sample the implausible triplets which are labeled as either entirely abstract or concrete, and convert them into sentences by inserting \enquote{the} before each subject and object.

\paragraph{CConS}
\citet{counter-common} creates the Counter-commonsense Contextual Size comparison dataset to investigate LLMs' commonsense reasoning capabilities at inferring relative size when given unusual sentence contexts.
Each sentence suggests uncommon relative sizes between two objects either explicitly (if the relative object sizes are explicitly stated), or implicitly (if  the verb implicitly suggests the relative object sizes).

\paragraph{Cusp}
\citet{metaphor-cusp} make 400 nonsense sentences by altering metaphors from \citet{Cardillo2010,Cardillo2016}. They 
select the top examples after either shuffling words across sentences, or running TextAttack's back-translation perturbation \citep{morris2020textattack}. 
Since the nonsense data is an extension of datasets previously published across multiple papers and was given no official name, we will refer to it as Cusp.

\section{Models}
By using LLMs, rather than embedding-based models, we are capable of analyzing model behavior across both tasks: contextualization and sensicality scoring of semantically deviant sentences. 
We select Microsoft's Phi 4 Mini Instruct \cite{phi} and Meta's Llama 3.1 8B Instruct \cite{llama} for their reported reliability across prompt variation \citep{weeber-etal-2026-political}.

The Phi 4 Mini base model has 3.84 billion parameters pre-trained on a 5 trillion corpus of web and synthetic data and post-trained on a series of NLP tasks including chain of thought (CoT) reasoning and summarization. 
Meta's Llama 3.1 is a base model of 8.03 billion parameters which are pre-trained trained on 15 trillion tokens of public-source data.
The Instruct versions we use are the result of fine-tuning for instruction prompting.

\section{Methods}

For each sentence in our dataset sample, we collect scores from both LLMs and humans following the process shown in Figure \ref{fig:methodology}.
We prompt both LLMs to generate: (a) a context where a given sentence could make sense; (b) a score for how sensical the sentence is without context; and (c)
 another score for the sentence within a generated context.
In parallel, we ask humans to annotate the sensicality of the sentences---with and without context.

In Section \ref{sec:human_ann}, we analyze if LLMs can hypothesize scenarios where unusual, seemingly nonsensical sentences are actually sensical by comparing the annotation scores for sentence sensicality with and without a generated context. Then in Section \ref{sec:can_llm_score}, by comparing the LLM-generated sensicality scores to the human annotations, we determine if LLMs are capable of sensicality scoring. Finally, Section \ref{sec:pattern_contexts} categorizes the generated contexts to analyze if the LLMs primarily utilize generic patterns or sentence-specific rationalizations.

\begin{figure*}[tb!]
  \centering
    \includegraphics[width=0.8\columnwidth]{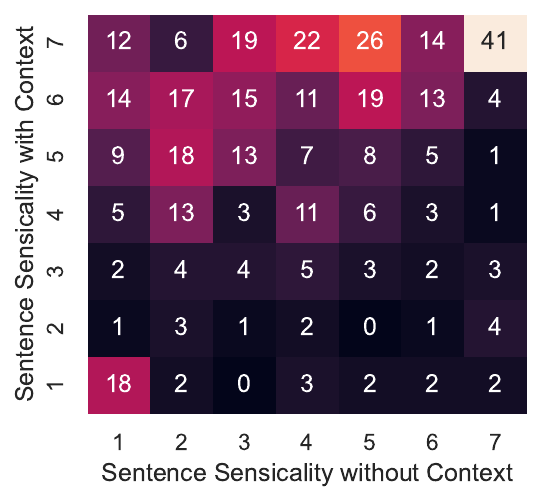} 
    \includegraphics[width=0.8\columnwidth]{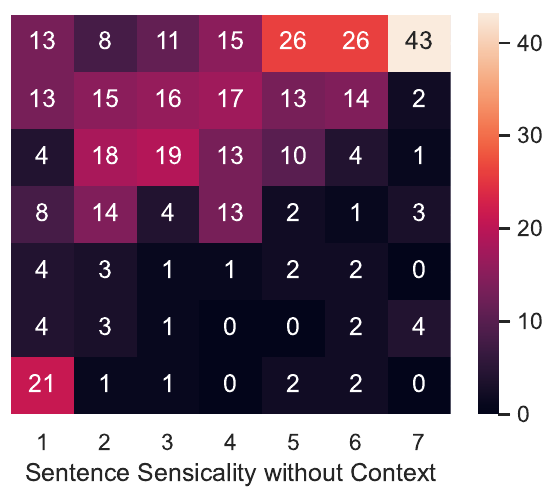} 
\caption{Effect of context on human sentence sensicality ratings (left: Phi context, right: Llama context)} \label{fig:ann_con_effect}
\end{figure*}

\subsection{LLM Prompting}\label{dev-prompt}

All prompting is run with the LLMs' default settings and a 128 token limit. Each prompt is chosen based on performance on a held-out dev set.  
During prompt development, requesting contexts to be \enquote{a few sentences long} yielded the most explanatory yet concise contexts.
The prompts used can be found in Table \ref{tab:prompts} in Appendix \ref{sec:study_layout}.

Following previous research, we judge sensicality as a graded measurement \citep{eva-first,spicy,ling-terminology}, and choose a 7-point Likert scale for sensicality, similar to \citet{howling-bert}, where 1 is \enquote{nonsense} and 7 is \enquote{makes sense.}
We generate scores for the sentences on their own, and when given a generated context.
Each prompt has two forms, asking the model to either \enquote{score} or \enquote{rate} the sentence, resulting in two scores per task per model.
Scores are also generated using a context from each LLM.
This means that each LLM scores each sentence six times: twice without a context, twice with its own context, and twice with the other LLM's context. 
Since both LLMs' score outputs include lengthy justifications, we developed a simple script to extract scores and manually reviewed ambiguous outputs. Details can be found in Appendix \ref{sec:get_score}.

\subsection{Human Annotations}

In order to determine if the LLM-generated contexts are comprehensible scenarios where semantically deviant sentences indeed make sense, we run an annotation study, asking two crowdsourcing workers to score a sentence's meaningfulness on its own and when considering a given LLM-generated context.
Each sentence is annotated four times: twice with a context generated by Phi, and twice with one by Llama.

Crowd-source annotations are run on Prolific by 16 workers, from a default population distribution who are fluent in English. We analyze two main questions:
How much does a given sentence make sense (1 as \enquote{Nonsensical,} 4 as \enquote{Neutral,} and 7 as \enquote{Makes Sense}), and then, after providing a generated context to the annotator, how much the sentence makes sense within the context (using the same scale as the first question). Figure~\ref{fig:annotation-example} in Appendix \ref{sec:study_layout} shows the annotation layout.
We check the robustness of the annotations by computing the Kendall tau-b correlation coefficient, recommended in the literature for evaluating Likert scale ratings \cite{agresti02}. Kendall's $\tau$ for both scoring tasks (with and without context) is almost identical at 0.164 ($p<0.001$) -- a highly significant but noisy relationship, as is often the case with lexical judgments. 
\changed{The subtypes differ in annotation robustness, though, more in the non-contextualized setting (average $\tau$=0.142, $\sigma$=0.136) than in the contextualized setting (average $\tau$=0.157, $\sigma$=0.05).}

We carry out all subsequent analyses on the raw scores for both annotators, unless stated otherwise. We make the annotations  available at: \changed{\url{https://github.com/KatrinaROlsen/NonsenseContext}}. 

\begin{figure*}[tb!]
    \includegraphics[width=0.50\textwidth]{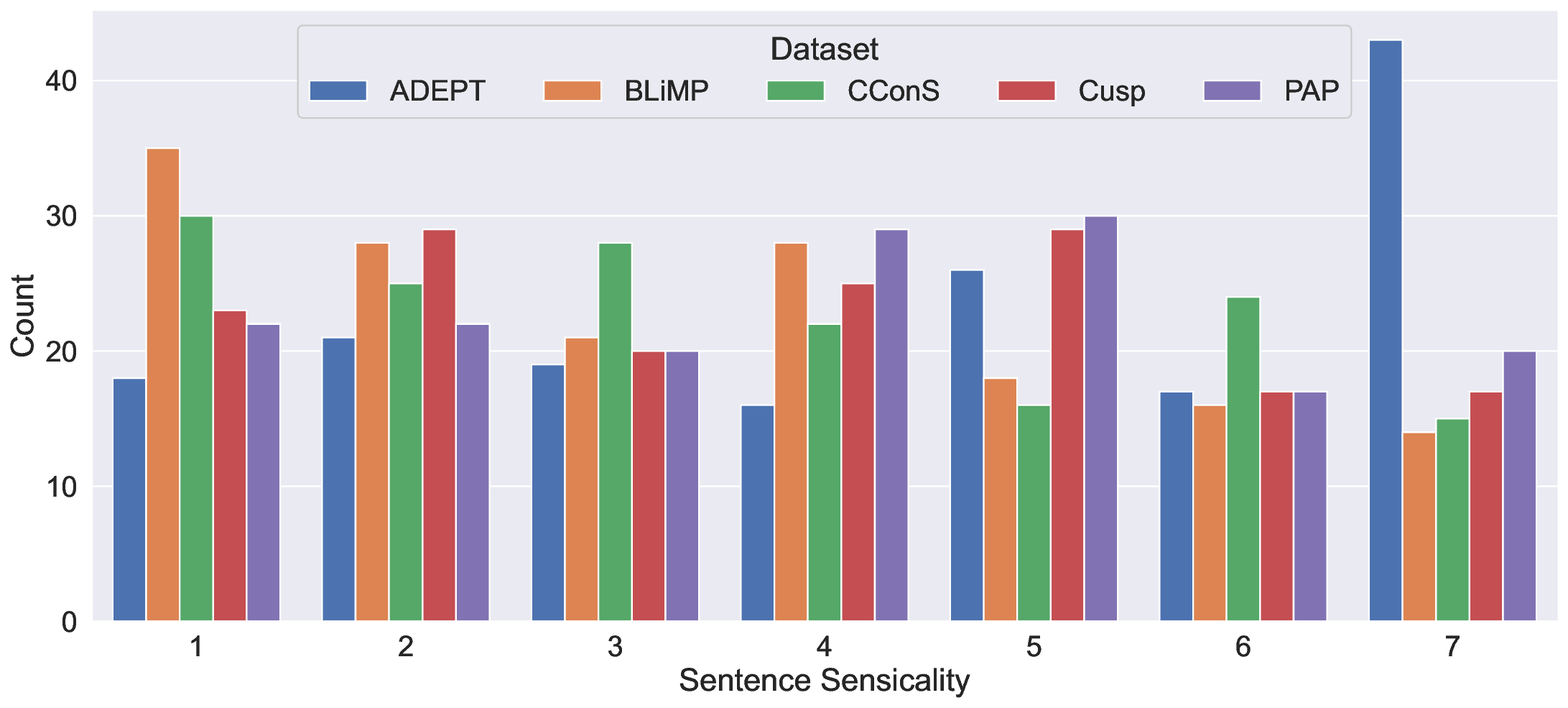}
  \hspace*{\fill}   
    \includegraphics[width=0.50\linewidth]{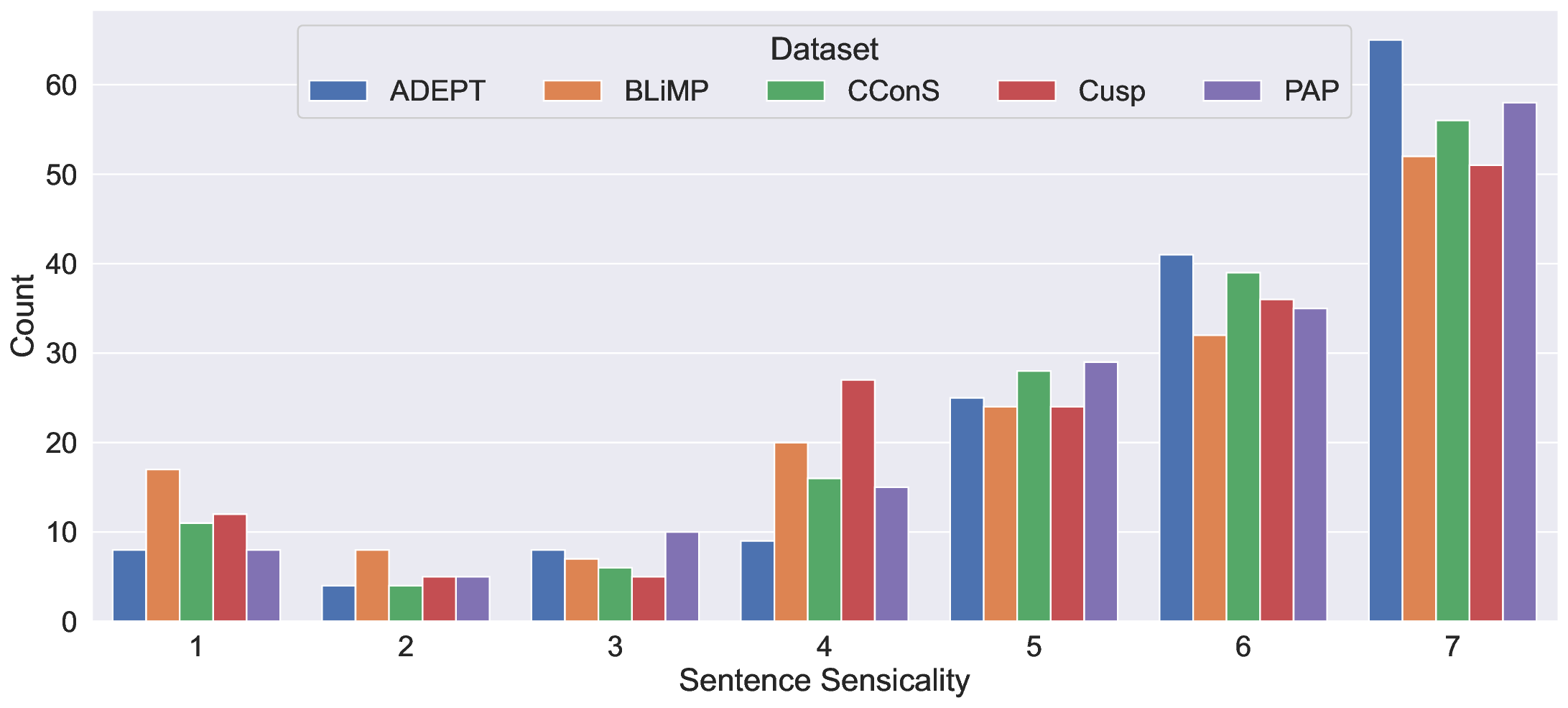}
  \caption{Human annotation sensicality scores by dataset (left: without context, right: with LLM-generated context)}
\label{fig:sensicality-by-datset}
\end{figure*}

\section{Can LLMs Contextualize Semantic Anomalies?}\label{sec:human_ann}

We now ask first how well LLMs can generate facilitating contexts for semantically anomalous sentences as the first aspect of RQ2. We answer this question by analyzing how the inclusion of a context changes how human annotators score a sentence's sensicality, relative to the sensicality of the sentence when annotated by human raters without context.
Figure \ref{fig:ann_con_effect} shows the results as count-based heat maps, where the x axis corresponds to sensicality without context, and the y axis sensicality in a context generated by Phi (left-hand panel) and Llama (right-hand panel).

For both models, the measured sensicality of the sentence increases in context, with the majority of scores residing in the upper left half of each figure. Numerically, the average meaningfulness of the sentences increases from 3.87 to 5.33 when involving contexts generated by Phi, and 3.85 to 5.38 when involving contexts generated by Llama.

While the figures appear very similar in how each LLMs' contexts affect sensicality, the Llama model produces slightly more contexts that are given the maximum sensicality score than Phi (140 for Phi, 142 for Llama), and slightly less contexts which are given fully nonsensical score (29 for Phi, 27 for Llama). The Phi model's contexts are able to increase the sensicality score of semantically deviant sentences 72.7\% of the time, and the Llama model's contexts 76.4\% of the time. Of the sentences whose contexts did not increase the sensicality score---94 for Phi, 83 for Llama---only 38 were not improved by either model, indicating that neither of these models have managed to 
reached a theoretical maximum performance on this dataset.

Examples of sentences that neither context was able to make sensical are \enquote{Every eggplant wasn't dropping by every mall.} and \enquote{The posterity conducts the origin.}
As a converse example, \enquote{The thunderstorm runs the portrait.} is rated by both annotators as 1, \enquote{Nonsensical,} and is made almost fully sensical (6) by both model's contexts: Phi's fantastical context where storms can animate objects and Llama's description of lightning flashing, making the objects within a portrait appear to run.

\subsection{Contextualizations by Dataset}

\begin{figure*}[tb!]
    \includegraphics[width=0.50\linewidth]{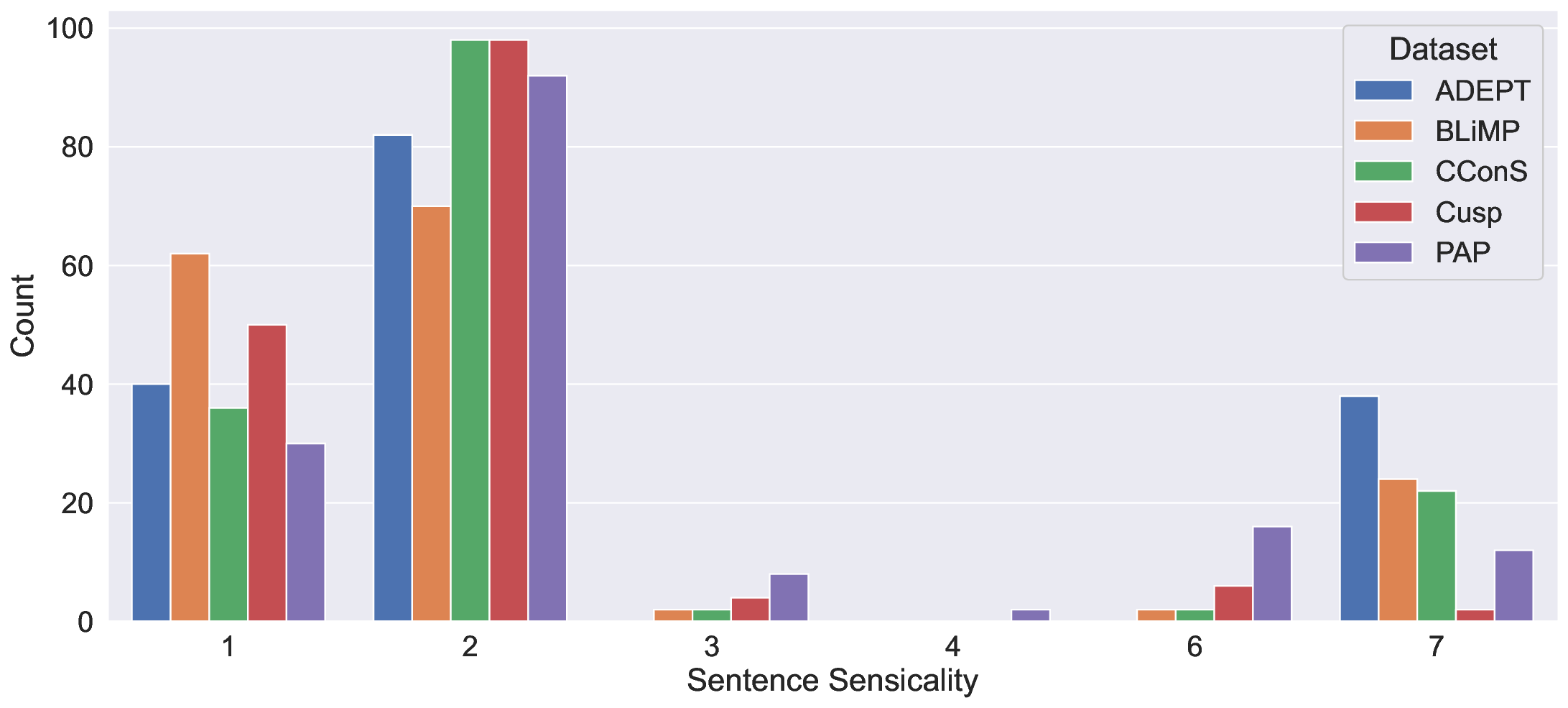}
  \hspace*{\fill}   
    \includegraphics[width=0.50\linewidth]{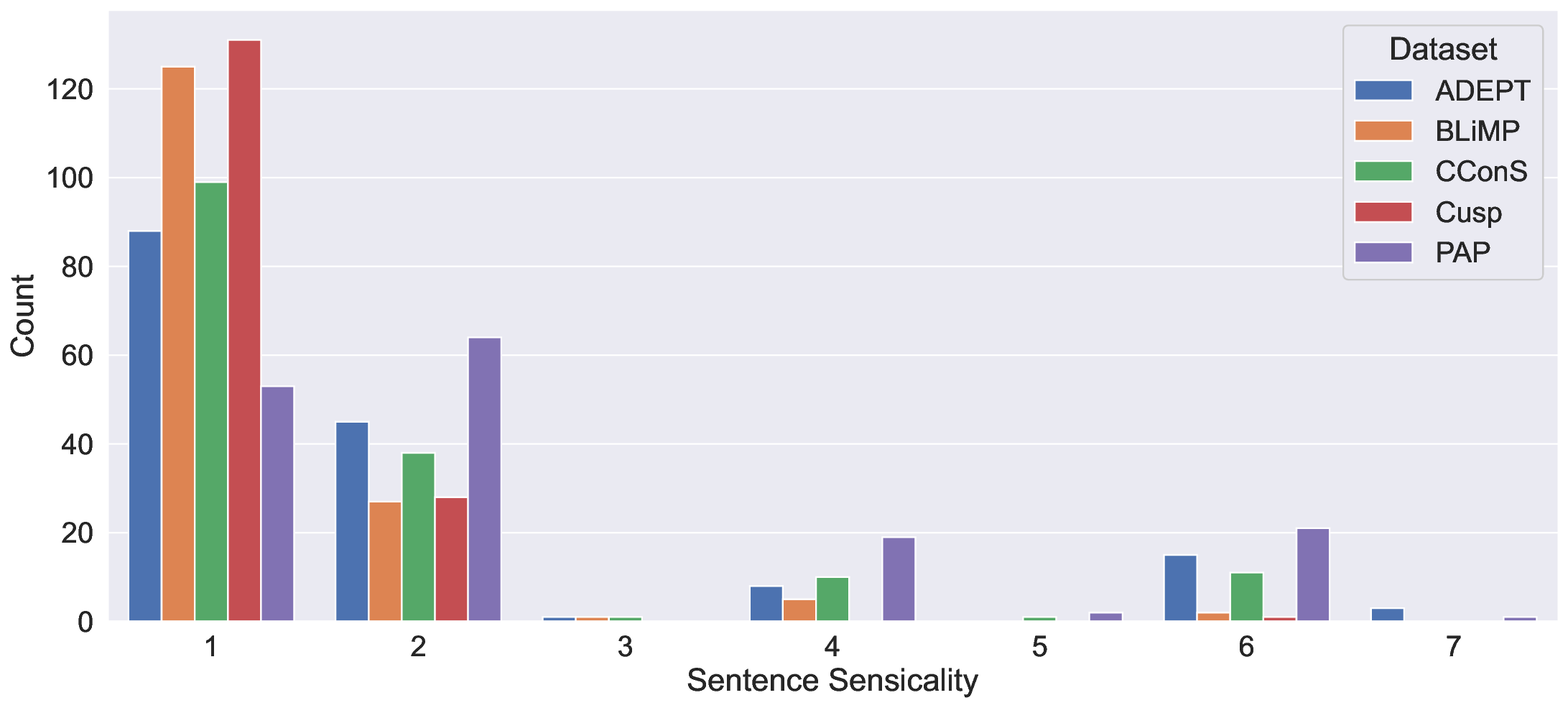}
  \caption{LLM scores on sentence sensicality by dataset (left: Phi, right: Llama)}
  \label{fig:sense_score_no_con}
\end{figure*}


The annotation scores for how much a sentence makes sense per dataset (RQ1) can be found in Figure \ref{fig:sensicality-by-datset} (left: without context, right: in context).
Immediately evident in the left-hand panel of Figure \ref{fig:sensicality-by-datset} is that the human annotators have not scored the majority of sentences as purely nonsensical, as one might expect for a collection of datasets which were made to be nonsensical, unacceptable, or impossible.
In fact, the overall counts for the sentences are almost level across all sensicality scores, with the counts only minorly decreasing as sensicality increases and score 6 being slightly less common than score 7. The BLiMP sentences are the closest representation to the outcome one would expect for nonsensical sentences, where the most common sensicality score is 1 and the least common is 7. ADEPT, on the other hand, has more than double of its sentences rated a 7 for sensicality than for any other score. In fact, 36\% of the ADEPT examples in the annotation dataset were given a 7 for meaningfulness by both annotators, including sentences like \enquote{An entangled mammal plays.} and \enquote{You use a false needle to draw a thread.} 
\changed{Such high sensicality ratings suggest that the adjectival modifications in ADEPT which yield a sentence to be impossible should not be considered as nonsensical in future research.}

Despite the overall level distribution of sensicality scores for the sentences in isolation, the effect of contextualization on sensicality scores is clear across all datasets in the right-hand panel of Figure \ref{fig:sensicality-by-datset}. While the ADEPT datasets are still slightly more often sensical than the others, and BLiMP remains to be the dataset with the most sentences scored at the lowest sensicality, the differences between datasets is lessened---rather than exasperated---with the inclusion of contexts, suggesting that the LLMs have no strong bias in their capability to contextualize one dataset over another. Indeed, the total range in the average amount that the sensicality score is raised per dataset is 0.335, with Cusp having lowest average increase in sensicality (1.741) and CConS having the largest increase on average (2.076).
This leveling effect in sensicality scores when given a context is also evident when viewing the data by subtypes (shown with Figure \ref{fig:subtypve_annotations} in Appendix \ref{app-subtype}).

\begin{figure}[tb!]
    \centering
    \includegraphics[width=\linewidth]{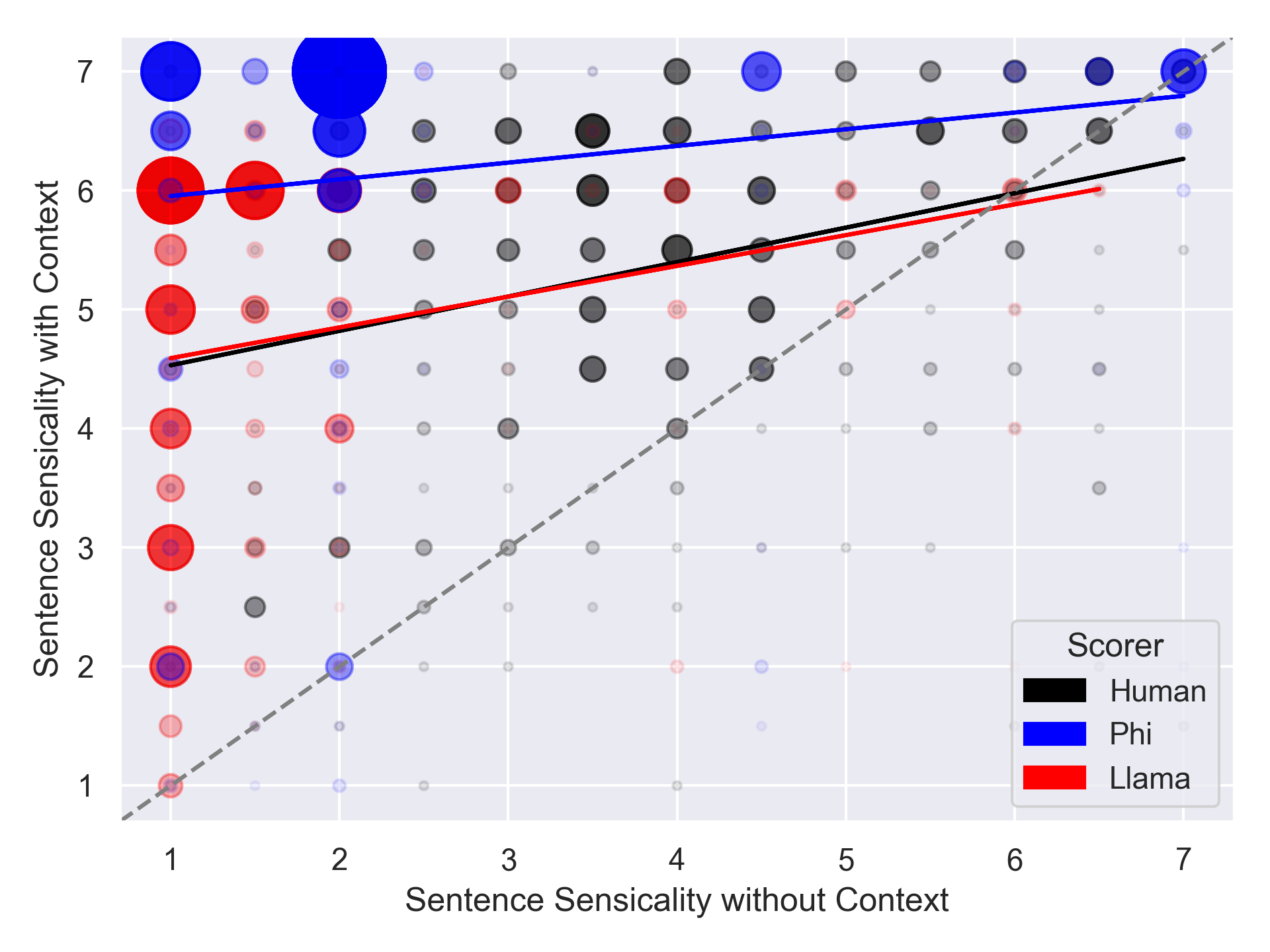}
    \caption{Effect of Context Sensicality Scoring by Humans and LLMs (dashed line: without context; solid lines: regression lines for scorers)}
    \label{fig:con_effect_score_human_models}
\end{figure}

Only 45.5\% of the semantically deviant sentences are annotated as purely nonsensical. This decreases further to 24.5\% in LLM-generated contexts. Beyond the demonstration of LLMs' capability of generating contexts that make unusual utterances meaningful, this also shows that many of the ``impossible'' and ``nonsensical'' examples from  the literature are in fact meaningful, even on their own.

\section{Can LLMs Score Sensicality?}\label{sec:can_llm_score}

We now move on to the second aspect of RQ2, asking  how consistently each LLM judges sensicality across datasets, how the scores vary from the humans' in Section \ref{sec:human_ann}, and whether either LLM has a bias when scoring on contexts it has generated.

\subsection{LLM Sensicality Scores}

The sensicality scores generated by Phi and Llama for our dataset collection are reported in Figure~\ref{fig:sense_score_no_con}.
\changed{Given the noted prompt variability, each model was prompted two times, resulting in four meaningfulness scores per sentence/context pair.}
Immediately evident is that both LLMs score all datasets as significantly less sensical than the human annotations. 
Llama most often scores sentences as completely nonsensical with the 1 label, with Phi's most common label being 2.
The Phi model also gives the highest sensicality score more often than Llama---for ADEPT in particular, which aligns with the human annotations. 

These lower scores are more in line with expected results on impossible and nonsensical data than the human annotation scores.
The LLM results also support the reasoning that impossible data would be more sensical than data specifically created to be nonsensical, with the relative lower sensicality for the BLiMP and Cusp datasets compared to the relatively higher scores for the plausibility datasets (ADEPT and PAP).

Given that human annotators alternated between sentence-only and with context judgments in our study, while LLMs were zero-shot prompted for each sentence, it is possible the human annotators were more cued to consider sentences within their own hypothetical contexts than the LLMs were, resulting in giving higher sensicality scores than LLMs.
In which case, LLMs more often giving sentences pure \enquote{nonsensical} scores would imply a lack of consideration for potential contexts when judging sentence sensicality---despite the demonstrated ability for both models to generate contexts which yield high sensicality scores.

\begin{figure*}[tb!]
    \includegraphics[width=0.50\linewidth]{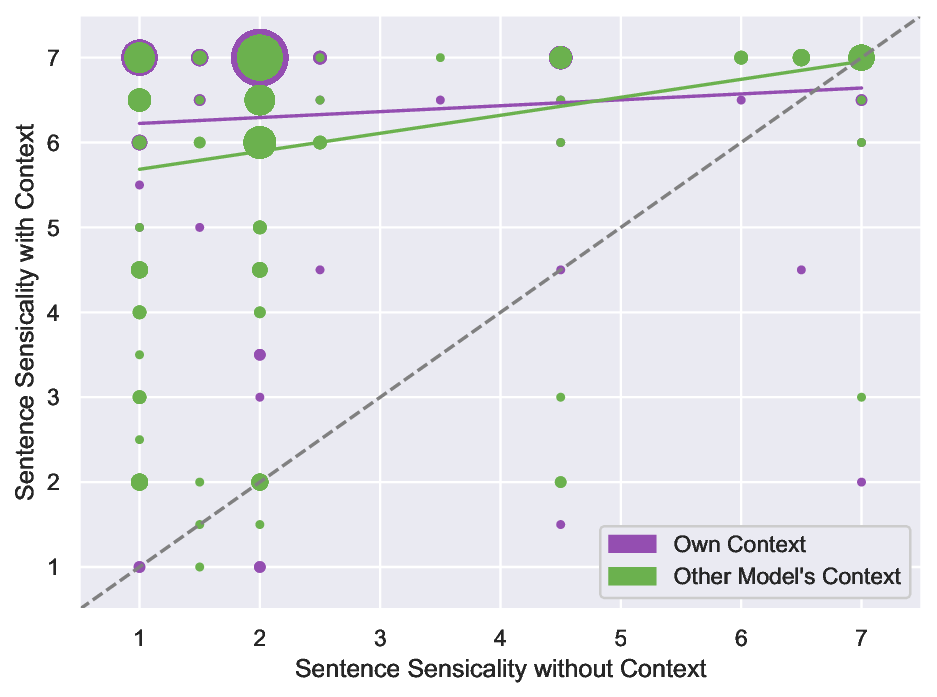}
  \hspace*{\fill}   
    \includegraphics[width=0.50\linewidth]{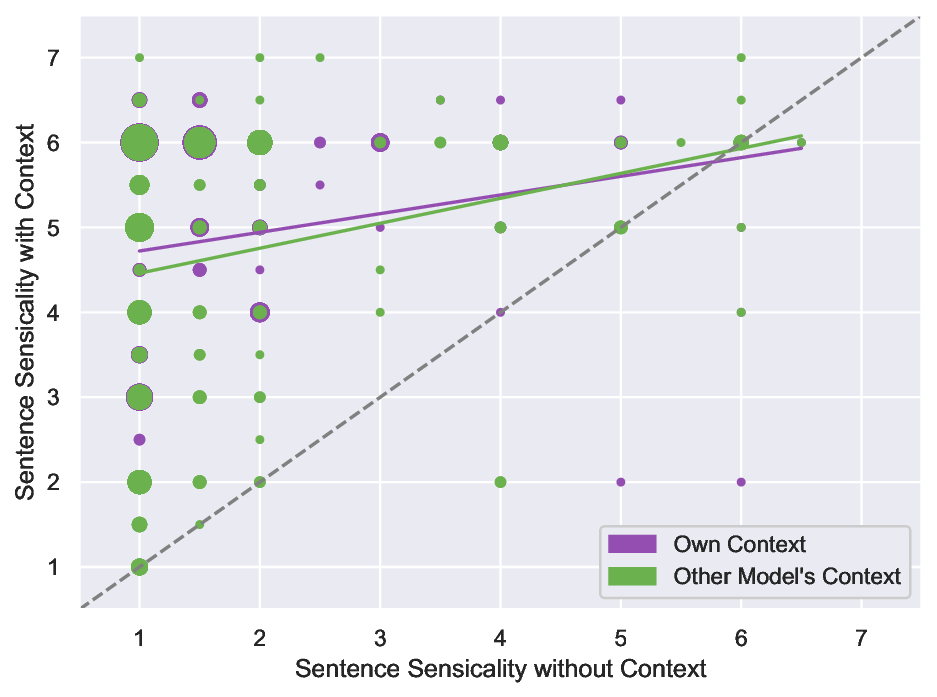}
  \caption{LLM creation sensicality bias on source of context (left: Phi scoring, right: Llama scoring). Solid lines are regression lines for different context sources.}
  \label{fig:bias}
\end{figure*}

\subsection{LLM Scores Compared to Humans}

Next we compare the effect of context on LLM sensicality scoring relative to the human annotation scores.
Following the evaluation of changes in acceptability when given a context in \citet{accpt-in-context}, the mean score between both human annotators per sentence and the mean score between both sensicality scoring prompts for both Phi and Llama are reported as scatter plots in Figure \ref{fig:con_effect_score_human_models}, where the color represents what gave the score and the dot size represents the number of sentences that are annotated as having any given level of sensicality, with and without context.
The dashed grey line denotes the line of equality between axes, and the full lines for each color represents the regression line for the scatter points of the same color. 
Both regressions are highly significant at $\alpha$=0.001.

Figure \ref{fig:con_effect_score_human_models} shows that, despite the increase of sensicality on nonsensical sentences, the inclusion of context also lowers the sensicality of fully sensical sentences---a common acceptability compression effect theorized by \citet{accpt-in-context} as due to human cognitive load. 
Interestingly, both models also experience the compression effect, with the Llama model's regression line almost identically matching the one for the human annotations.
Although the Phi regression line has a similar slope for increased sensicality when given a context, the Phi model more often gives fully sensical scores when given a context (18 times more often than Llama), resulting in a regression line that is much higher than the human and Llama lines.
Although both LLMs portray quite stricter sensicality judgments for sentences on their own,
the similarity of the LLM regression lines to the human's, especially for the Llama model, suggests that
LLMs experience the relative effect of context on the perceived sensicality of a sentence quite similar to humans.
Future research would be enlightening as to if this similarity in context effect extends to a wider range of sensical sentences.

\subsection{Scoring  Bias for Own Generated Contexts}

Since LLMs are scoring relative to their own generated contexts half the time, it is worth evaluating if either model has a bias for its own generated contexts. The left-hand panel of Figure \ref{fig:bias} demonstrates how Phi scores the sentence sensicality when given its own generated context and when given one made by Llama; and the right-hand panel shows the reverse, having Llama's scores on its own contexts versus Phi's. 
Phi's tendency to score high sensicality is not exclusive to its own contexts; however, it also is revealed that Phi's scores when given its own context remain almost constant, no matter how sensical Phi scored the sentence without a context.

For the most sensical sentences, both models give higher scores for the context they did not generate than for their own. 
However, given the small number of sentences rated as sensical in Figure \ref{fig:sense_score_no_con}, it is difficult to draw strong conclusions.

Overall, we demonstrate that, although both LLMs score sentence sensicality drastically different than humans, the effect of including a context when scoring sensicality is quite similar between LLMs and humans, and that there is little bias for models scoring on own-generated contexts.

\begin{table*}[tb!]
\begin{tabular}{
    p{1.5cm}lll  p{3.5cm}p{6.1cm}
  }
  \toprule
\textbf{Theme} & \textbf{Total} & \textbf{Llama} & \textbf{Phi} & \textbf{Example Sentence} & \textbf{Example Generated Context}   
\\ \midrule

Specific Setting & 171            & 102            & 69           & The collaboration facilitates the ambience.                            & In a corporate event setting, the collaboration between the event planners...                                              \\
\rowcolor[HTML]{EFEFEF} 
Fantasy          & 78             & 31             & 47           & A human is evacuated from a bottle.                                    & A sensical context for \enquote{A human is evacuated from a bottle} could be a science fiction scenario where...                   \\
Figurative       & 55             & 25             & 30           & The therapy was an archeological laughter.                             & In this context, \enquote{The therapy was an archeological laughter} is a metaphorical phrase suggesting...                        \\
\rowcolor[HTML]{EFEFEF} 
Story            & 41             & 16             & 25           & \cellcolor[HTML]{EFEFEF}Some cake was talking about Brian.             & \cellcolor[HTML]{EFEFEF}In a whimsical, fantastical story, imagine a magical world where inanimate objects come to life... \\
Humorous         & 39             & 21             & 18           & The school was run around by some muffin.                              & In a whimsical or humorous context, \enquote{The school was run around by some muffin} could make sense if...                    \\
\rowcolor[HTML]{EFEFEF} 
Art              & 37             & 19             & 18           & A human watched a salmon disappear into a zipper.                      & A human watched a salmon disappear into a zipper during a unique art installation...                                       \\
Children         & 24             & 8              & 16           & A bear was displayed in a teddy bear.                                  & In a children's museum or an exhibit on taxidermy, a real bear might be displayed...                                       \\
\rowcolor[HTML]{EFEFEF} 
Jargon           & 24             & 20             & 4            & Small bush wins no more.                                               & The phrase \enquote{Small bush wins no more} is a colloquialism often associated with the game of poker...                        \\ \bottomrule
\end{tabular}
\caption{Common themes in generated contexts}
    \label{tab:gen_con_themes}
\end{table*}

\section{Qualitative Characteristics of Generated Contexts}\label{sec:pattern_contexts}

When evaluating how well LLMs can hypothesize sensical contexts for seemingly nonsensical sentences, it is important to understand not just the effect of contexts on sensicality annotations, but also the specificity of the contextualizations for the given sentences---whether the LLMs are generating contexts specific to the sentences given, or whether they often use generic settings, such as a fantasy story, which could apply to almost any sentence.
To assess the extent to which our LLMs rely on specific narrative patterns for explanations, we label each generated context for sense-making themes.

\subsection{Context Patterns by LLM}

We label themes according to the model's direct self-labeling statements in the contexts, rather than through textual interpretation. 
This is largely due to self-contradictions in generated text, such as a context stating its interpretation of a sentence is both literal and metaphorical. We \changed{manually} label contexts by both themes instead of guessing which the model \enquote{intended.}
We determine each theme by patterns of keywords in contextualizations, e.g., the Story theme for if the context says a sentence makes sense in a \enquote{story,} \enquote{novel,} \enquote{book} etc.
The average number of labels per context is 1.2.

The most common themes in the generated contexts are reported in Table \ref{tab:gen_con_themes}, along with the instance counts for each LLM's contexts. The full example contexts are given in Table \ref{tab:full_gen_con_themes} in the Appendix.

\begin{table*}[tb!]
\centering
\begin{tabular}{clllll}
\toprule


\textbf{Theme} & \textbf{ADEPT} & \textbf{BLiMP} & \textbf{PAP} & \textbf{CConS} & \textbf{Cusp} 
\\ \midrule

Specif. Setting & 36             & 30             & 47           & 31             & 27            \\
\rowcolor[HTML]{EFEFEF} 
Fantasy          & 13             & 19             & 14           & 17             & 15            \\
Figurative       & 9              & 7              & 9            & 4              & 26            \\
\rowcolor[HTML]{EFEFEF} 
Story            & 5              & 8              & 6            & 12             & 10            \\
Humorous         & 3              & 10             & 3            & 9              & 14            \\
\rowcolor[HTML]{EFEFEF} 
Art              & 6              & 4              & 0            & 25             & 2             \\
Children         & 1              & 2              & 2            & 15             & 4             \\
\rowcolor[HTML]{EFEFEF} 
Jargon           & 11             & 3              & 1            & 4              & 5            \\ \bottomrule
\end{tabular}
\caption{Frequency of generated themes in datasets}
    \label{tab:gen_con_themes_per_origin}
\end{table*}

Some themes are unsurprisingly common, such as stating the sentence is in a fantasy setting, in a story, or relating to children (e.g., their games, toys, etc.). 
While these three themes often co-exist in the contexts, many non-fantastical, non-children stories are present in our data. 
The Art theme is when a sentence is stated to occur because physical art is being performed---often as art installations or on stage in a play. 
The Humorous theme refers to when a model labels a context as comedic or humorous; 90\% of the contexts with this theme contain the exact word \enquote{humorous.} 
While perhaps surprising as a device to contextualize semantically deviant sentences---calling something humorous doesn't necessarily make it less nonsensical---the prevalence of this theme may be because jokes and silly events have a lower threshold for believability. 

Notably, the Fantasy, Story, and Figurative themes are less common than Specific Setting, which is where the context states a unique setting in which the specific contents of the nonsensical sentence would make sense. For example, a generic contextualization for \enquote{The moon revolves the album.} would say that a miniature moon is rolling around an album; whereas a Specific Setting example would establish \enquote{in the context of astronomy} or \enquote{when in a music production setting} to justify why and how the moon would revolve the album.

The least common theme, Jargon, argues that the input sentence has an alternate meaning when used in a specific linguistic context. These are stated as facts, although many of the examples, such as the poker phrase stated in the Jargon example in Table \ref{tab:gen_con_themes}, have no instances online as of writing.

\subsection{Context Patterns by Dataset}

Returning to RQ1, we observe that although the Llama model more often has a Specific Setting and Jargon in its contexts, and Phi has more Fantasy instances, neither model depends on a single theme for explanations more than the other.
Larger theme imbalances are, however, evident by dataset, shown in Table \ref{tab:gen_con_themes_per_origin}, particularly with Art and Children for CConS, and Figurative for the Cusp dataset.

The contexts explaining examples from the CConS dataset, which focuses on uncommon relative sizes, take up over half of the total instances of the Art and Children theme. This may be because both themes allow for easy explanations of relative sizes, such as by stating that a giant light bulb has been made to contain a fan for an art installation, or that a child's toy ship sinks in their cereal bowl.

Next, the Cusp dataset is contextualized by using figurative language two to five times more often than the other datasets in our sample dataset. 
This is intriguing since, although no valid metaphors are present in the nonsensical sentences themselves, the Cusp nonsensical data is the only dataset originally created from metaphoric sentences rather than from physically plausible sentences. 
It is possible the original metaphoric vocabulary or syntax structures remaining in Cusp's nonsensical data could be cueing figurative usage to the LLMs. Or the vocabulary from the original metaphoric sentences, which can be abstract and involve complex cognitive frames, could happen to most naturally be explained via figurative language.

Finally, the Fantasy and Specific Setting context themes have the most consistent counts across datasets.
This may be due to the previously stated flexibility of the Fantasy theme to explain almost any unusual sentence and the broad nature of the Specific Setting theme.

\section{Conclusion}
 
Using LLM-generated contexts and a human annotation study, we answer our first research question by finding potential meaning in many of the sentences previous research has labeled as impossible, semantically unacceptable, and nonsensical---both on their own and with a context.
Such unexpected meaningful interpretations may be a warning bell that abstract contextless judgments on a sentence's semantics are not fully representative of the sentence's semantic potential.
It is worth evaluating in future research if models trained on this data are subsequently overly strict on unusual but acceptable, real world data.

Regarding our second research question, we demonstrate the capability of both tested LLMs to generate hypothetical situations where unusual sentences are sensical, with Phi and Llama's contexts increasing sentence sensicality 72.7\% and 76.4\% of the time, respectively.
Both LLMs' contextualizations tend to be rather specific to the given unusual sentence, with the strongest patterns being found within datasets rather than across datasets.
The contexts' success in raising sensicality shows that both LLMs can compositionally utilize meanings learned from their sensical, pre-training data in order to parse potential meanings in unusual or nonsensical sentences.
However, a hypothetical maximum performance has not been reached,
given that both models generate some contexts which do not raise the sentences' sensicality as much as its counterpart's contexts.

Given \changed{the demonstrated effect that context has} on semantic interpretation and our reported high sensicality ratings on data previously considered meaningless, we encourage a broader use of semantic acceptability rankings \textit{in context} in future research.

\section{Limitations}
Since the datasets we use are publicly available, we recognize the possibility of their presence in the LLMs' pre-training data.
However, because all the pre-existing examples were artificially generated and have no original context for models to refer to, we find the effect of this possibility minimal on the LLMs' context generation performance.

In order to provide in-depth analysis of the generated contexts, we analyze the results of only two LLMs. Testing a wider array of LLMs would be insightful for how task performance can relate to architectures and training methods. \changed{We similarly believe a larger number of annotators would be insightful such creative, subjective tasks.}
Additionally, we consider this important work to perform across languages, not just English, and encourage more multilingual nonsense research in the future.

\bibliography{custom}

\appendix

\section{Study Details}\label{sec:study_layout}

\changed{In our human study, for each sentence/context pair we ask a total of five required questions, answered on a 7-point Likert scale. First, we provide a sentence and ask: (1) \enquote{How much does this sentence make sense?} (ranging from 1 Nonsensical, to 4 Neutral, to 7 Makes sense) and (2) \enquote{How easy is it to think of a context where the sentence would make sense?} (ranging from 1 Very Difficult, to 4 Neutral, to 7 Very Easy).}

\changed{Next we provide a LLM-generated context. Based off this we ask three more Likert-scale questions: (3) \enquote{How much does the sentence make sense when considered within the given context?} (ranging from 1 Nonsensical, to 4 Neutral, to 7 Makes sense), (4) \enquote{How much of the sentence's meaning is explained by the context?} (ranging from 1 None of the Sentence, to 4 Parts, to 7 Entire Sentence), and (5) \enquote{How confident are you in your judgments for this section?} (ranging from 1 Very Unsure, to 4 Neutral, to 7 Very Confident). Finally, we provide an optional text field for the annotator to explain their reasoning when judging the sentence/context pair.}

\changed{The questions were chosen based on a preliminary analysis of a sentences/contexts in a held-out development set. Given the high correlation between questions 1 and 2, and questions 3 and 4, we only discuss the results of questions 1 and 3 in this paper. However, all data is made available at: \url{https://github.com/KatrinaROlsen/NonsenseContext}}

\begin{table}[h]
\resizebox{\columnwidth}{!}{%
\begin{tabular}{c|p{0.95\linewidth}}
\rowcolor[HTML]{C0C0C0} 
\textbf{Generation Type}                                                                        & \multicolumn{1}{c}{\cellcolor[HTML]{C0C0C0}\textbf{Prompt}}                                                                                                                                                                                                                                                                                                                                                                     \\
\rowcolor[HTML]{EFEFEF} 
\textbf{Context Generation}                                                                     & Describe in a few sentences a context where "{[}SENT{]}" is sensical.                                                                                                                                                                                                                                                                                                                                                           \\
\textbf{Sentence Sensicality Scoring}                                                           & {[}Score/Rate{]} the following sentence between 1 and 7 for how sensical it is: {[}SENT{]}                                                                                                                                                                                                                                                                                                                                      \\
\rowcolor[HTML]{EFEFEF} 
\textbf{\begin{tabular}[c]{@{}c@{}}Sentence Sensicality Scoring\\ Given a Context\end{tabular}} & {[}Score/Rate{]} the sentence "{[}SENT{]}" between 1 and 7 for how sensical it is given the context: {[}CONTEXT{]}                                                                                                                                                                                                                                                                                                                                                                                                                                                                                                                              
\end{tabular}
}
\caption{Prompts Used for Text Generation}
    \label{tab:prompts}
\end{table}

\begin{figure}[h]
    \centering
    \includegraphics[width=\linewidth]{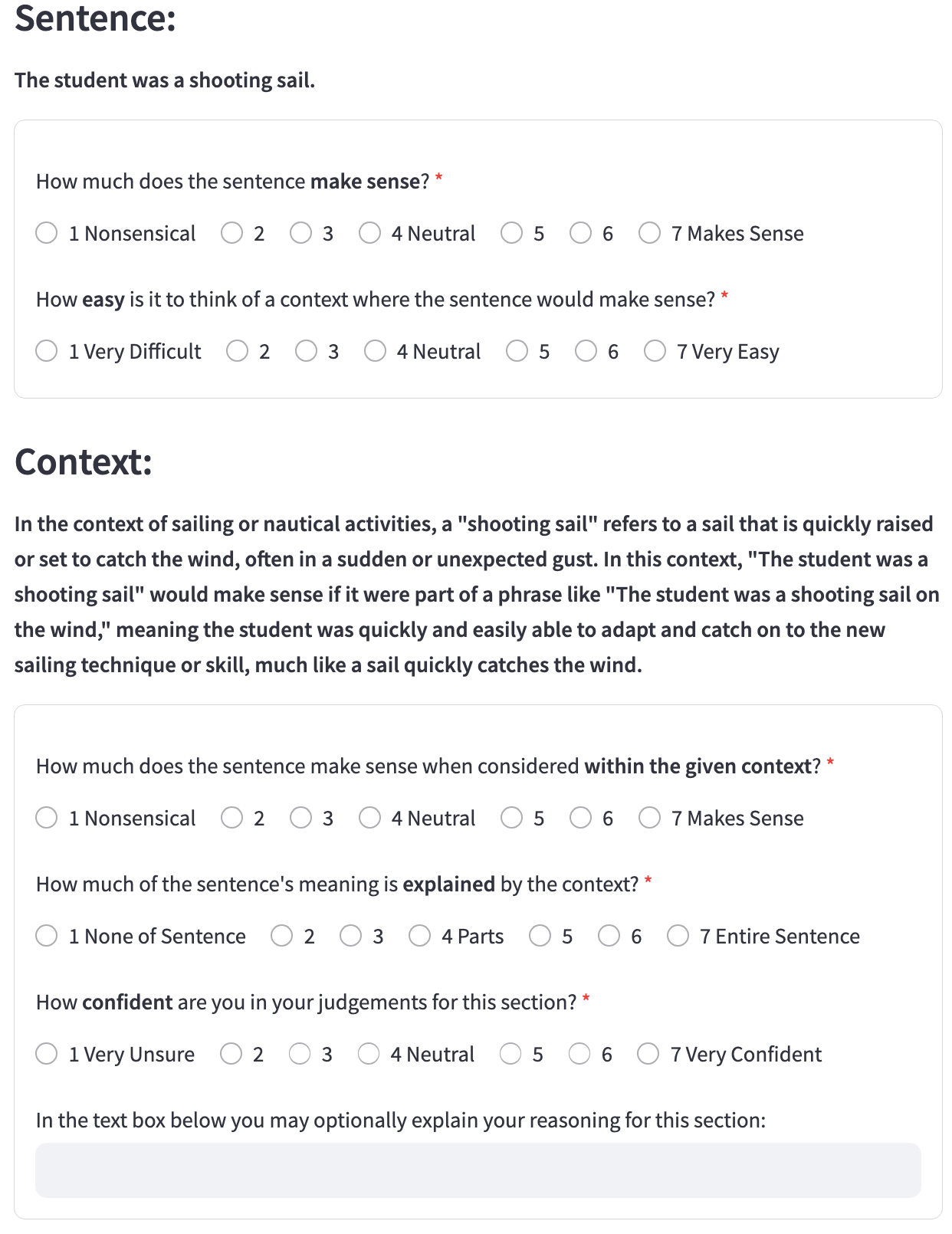}
    \caption{Annotation Example}
    \label{fig:annotation-example}
\end{figure}

\begin{table*}[h]
\resizebox{\linewidth}{!}{%
\begin{tabular}{c|c|c|c|p{0.5\linewidth}|p{0.9\linewidth}}
\rowcolor[HTML]{C0C0C0} 
\textbf{Theme}            & \textbf{Total} & \textbf{Llama} & \textbf{Phi} & \multicolumn{1}{c|}{\cellcolor[HTML]{C0C0C0}\textbf{Example Sentence}} & \multicolumn{1}{c}{\cellcolor[HTML]{C0C0C0}\textbf{Example Context}}                                                                                                                                                                                                                                                                                                                                                                                                                                                                                                                                          \\ \hline
\textbf{Specific Setting} & 171            & 102            & 69           & The collaboration facilitates the ambience.                            & In a corporate event setting, the collaboration between the event planners, catering staff, and audio-visual technicians is crucial. Their coordinated efforts ensure that the ambiance is perfect, with the right lighting, music, and refreshments, creating a welcoming and professional atmosphere for the attendees.                                                                                                                                                                                                                                                                                     \\
\rowcolor[HTML]{EFEFEF} 
\textbf{Fantasy}          & 78             & 31             & 47           & A human is evacuated from a bottle.                                    & A sensical context for \enquote{A human is evacuated from a bottle} could be a science fiction scenario where a person is trapped in a malfunctioning, airtight bottle due to a catastrophic event. Advanced technology or a rescue team might then use specialized equipment to safely extract the person from the bottle, ensuring their survival and well-being.                                                                                                                                                                                                                                                   \\
\textbf{Figurative}       & 55             & 25             & 30           & The therapy was an archeological laughter.                             & In this context, \enquote{The therapy was an archeological laughter} is a metaphorical phrase suggesting that the laughter that arose during therapy was a discovery or excavation of long-buried emotions or memories, much like an archaeologist uncovers ancient artifacts. The laughter in this case is not just a superficial response but a deeper, cathartic release of pent-up feelings, allowing the individual to confront and process their emotional past.                                                                                                                                                \\
\rowcolor[HTML]{EFEFEF} 
\textbf{Story}            & 41             & 16             & 25           & Some cake was talking about Brian.                                     & In a whimsical, fantastical story, imagine a magical world where inanimate objects come to life. In this world, a group of enchanted cakes, each with their own personalities, gather in a grand, enchanted kitchen. One of the cakes, perhaps a wise and ancient sponge cake, starts a conversation about Brian, a beloved character in the story who is known for his bravery and kindness. The cakes discuss Brian's adventures, his heroic deeds, and how he has positively impacted their world, making the phrase \enquote{Some cake was talking about Brian} a delightful and fitting part of the narrative. \\
\textbf{Humorous}         & 39             & 21             & 18           & The school was run around by some muffin.                              & In a whimsical or humorous context, \enquote{The school was run around by some muffin} could make sense if it's a fantasy world where a magical muffin has come to life and is now the de facto leader of the school. Perhaps the muffin has a charismatic personality, clever ideas, and a knack for getting everyone to follow its instructions, despite being a baked good. In this world, the muffin's authority is accepted by the students and staff, and it's not uncommon to see the muffin making announcements, handing out assignments, or even leading assemblies.                                      \\
\rowcolor[HTML]{EFEFEF} 
\textbf{Art}              & 37             & 19             & 18           & A human watched a salmon disappear into a zipper.                      & A human watched a salmon disappear into a zipper during a unique art installation at a local aquarium. The exhibit featured a large, transparent zipper embedded in a tank, and visitors were invited to release live salmon into the water. As the fish swam towards the zipper, they would occasionally swim into it, creating a surreal and captivating visual effect. The human observer was fascinated by the unexpected and whimsical interaction between the fish and the zipper, marveling at the salmon's ability to navigate the unusual obstacle.                                                  \\
\textbf{Children}         & 24             & 8              & 16           & A bear was displayed in a teddy bear.                                  & In a children's museum or an exhibit on taxidermy, a real bear might be displayed in a glass case or enclosure that resembles a giant teddy bear. The bear is then surrounded by other exhibits or interactive elements that create a whimsical, oversized teddy bear environment. This display would be an educational and visually striking way to showcase the bear, while also incorporating a popular children's toy as a creative framework.                                                                                                                                                            \\
\rowcolor[HTML]{EFEFEF} 
\textbf{Jargon}           & 24             & 20             & 4            & Small bush wins no more.                                               & The phrase \enquote{Small bush wins no more} is a colloquialism often associated with the game of poker. In this context, it means that a small or weak hand is no longer going to win the game, as the stakes are too high or the competition is too tough. This phrase is usually used to caution players against holding onto weak hands in a game where only the strongest hands will succeed.                                                                                                                                                                                                                   
\end{tabular}
}
\caption{Full Contexts by Common Themes}
    \label{tab:full_gen_con_themes}
\end{table*}

\section{Retrieving Scores from LLM Output}\label{sec:get_score}

Our score extraction script collects scores between 1 and 7 in generated text. Outputs were reviewed manually if the script found: no numbers, multiple numbers, or numbers that aren't the between 1 and 7. Of the flagged results, outputs which do not contain a single integer score within the provided scale are rejected from this paper's analysis. 

Both models provide a score for all sentences when no context was given.
When given a context, Llama declines giving a score twice. 
Phi declines twice, gives an invalid float score six times, and provides no score at all twice. For these cases where a score is missing, the score in Figure \ref{fig:con_effect_score_human_models} is the result of the alternate prompt, rather than the average between the two prompt variants.

\section{Sensicality Scores by Subtype}\label{app-subtype}

\begin{figure*}[h]
  \begin{subfigure}{0.5\textwidth}
    \includegraphics[width=\linewidth]{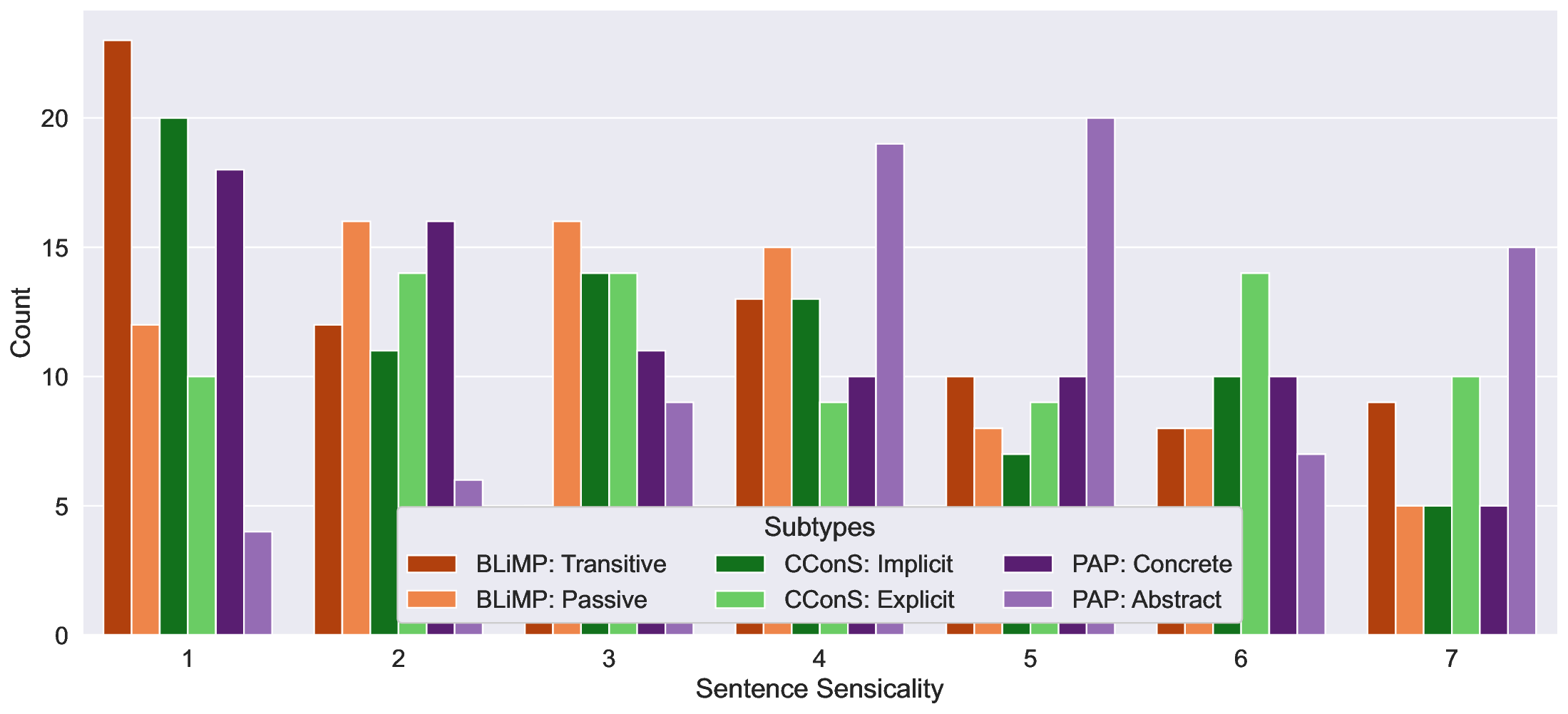}
    \caption{Sentence Meaning Without Context}
    \label{fig:annotate_no_con_p_type}
  \end{subfigure}%
  \hspace*{\fill}   
  \begin{subfigure}{0.5\textwidth}
    \includegraphics[width=\linewidth]{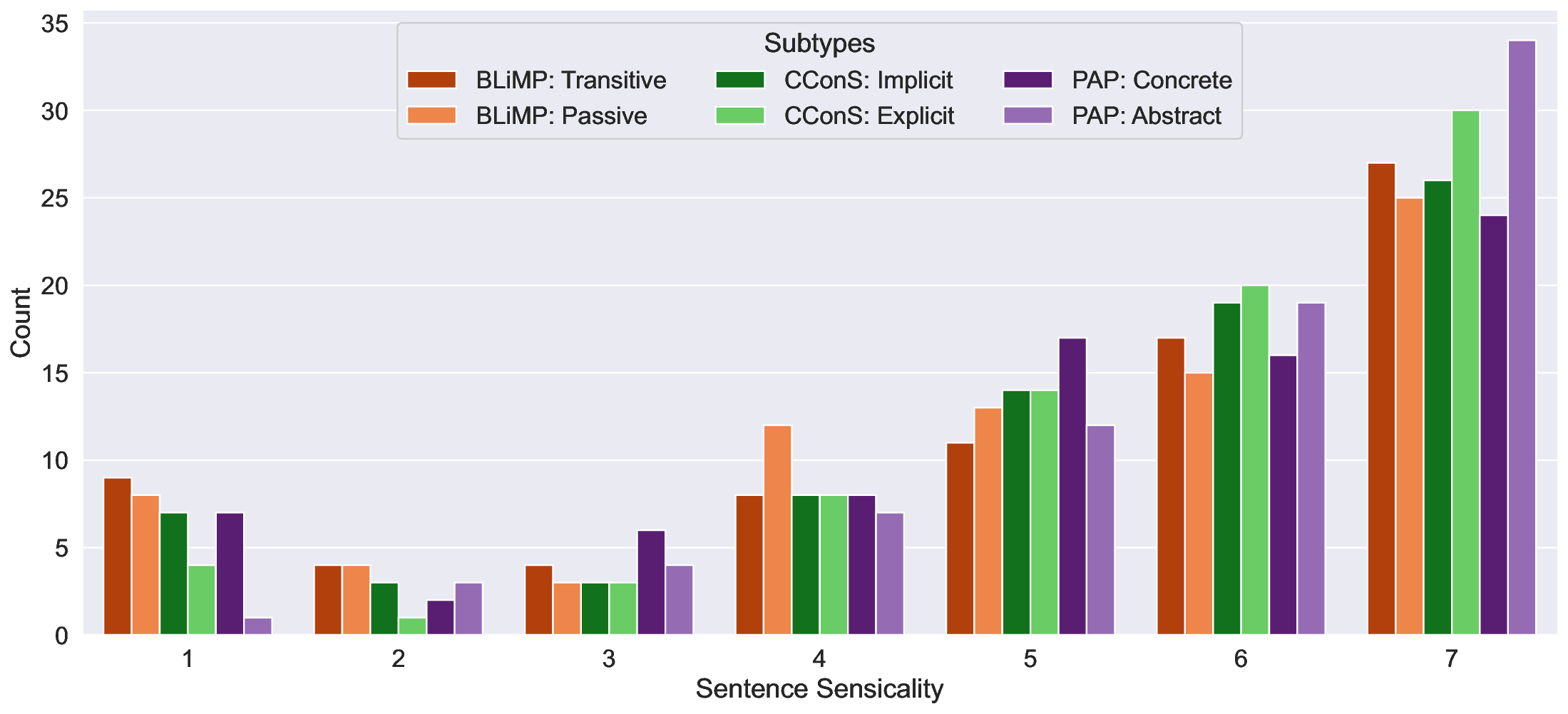}
    \caption{Sentence Meaning With Context}
    \label{fig:annotate_w_con_p_type}
  \end{subfigure}%
  \caption{Annotation Scores by Sub-Type}
  \label{fig:subtypve_annotations}
  \hspace*{\fill}
\end{figure*}

The human annotated sensicality scores are shown by each data subtype in Figure \ref{fig:annotate_no_con_p_type} for without context and in Figure \ref{fig:annotate_w_con_p_type} for with context.

\end{document}